\pdfoutput=1
\RequirePackage{fix-cm}
\documentclass[twocolumn]{svjour3}

\smartqed
\usepackage{graphicx}
\usepackage{amsmath,amssymb}
\usepackage{hyperref}

\usepackage{tikz}
\usetikzlibrary{arrows,arrows.meta,backgrounds,calc,patterns,positioning,shapes,shadows}
\usepackage[linesnumbered, ruled, noend]{algorithm2e}

\makeatletter
	\def\cl@chapter{} 
\makeatother

\usepackage{csquotes}
\usepackage{subcaption}
\usepackage[capitalise]{cleveref}
\usepackage[nolist, nohyperlinks]{acronym}
\usepackage{booktabs}
\acrodef{dag}[DAG]{directed acyclic graph}
\acrodef{dss}[DSS]{decision support system}
\acrodef{fmea}[FMEA]{failure mode and effects analysis}
\acrodef{llm}[LLM]{large language model}
\acrodef{mdp}[MDP]{Markov decision process}
\acrodefplural{mdp}[MDPs]{Markov decision processes}

\newcommand{\abs}[1]{\lvert #1 \rvert}
\newcommand{\causes}{\ensuremath{\mathrm{causes}}}
\newcommand{\detect}{\ensuremath{\mathrm{det}}}
\newcommand{\effects}{\ensuremath{\mathrm{effects}}}
\newcommand{\lc}{\ensuremath{\mathrm{left\_critical}}}
\newcommand{\occ}{\ensuremath{\mathrm{occ}}}
\newcommand{\poss}{\ensuremath{\mathrm{poss}}}
\newcommand{\prob}{\ensuremath{\mathrm{prob}}}
\newcommand{\range}{\ensuremath{\mathrm{range}}}
\newcommand{\rc}{\ensuremath{\mathrm{right\_critical}}}
\newcommand{\risk}{\ensuremath{\mathrm{risk}}}
\newcommand{\sev}{\ensuremath{\mathrm{sev}}}
\newcommand{\sign}{\ensuremath{\mathrm{sign}}}
\newcommand{\tooLow}{\ensuremath{\mathrm{tooLow}}}
\newcommand{\normal}{\ensuremath{\mathrm{normal}}}
\newcommand{\tooHigh}{\ensuremath{\mathrm{tooHigh}}}
\newcommand{\Ch}{\ensuremath{\mathrm{Ch}}}
\newcommand{\Pa}{\ensuremath{\mathrm{Pa}}}

\definecolor{myblue}{RGB}{0, 153, 255}
\definecolor{mybluedark}{RGB}{0, 51, 255}
\definecolor{mygray}{RGB}{153, 153, 136}
\definecolor{mygreen}{RGB}{0, 153, 51}
\definecolor{mypink}{RGB}{204, 51, 255}
\definecolor{mypurple}{RGB}{102, 0, 204}
\definecolor{myred}{RGB}{204, 0, 0}
\definecolor{myyellow}{RGB}{255, 204, 0}

\definecolor{newblue}{RGB}{50,113,173}
\definecolor{newred}{RGB}{222,32,36}
\definecolor{newgreen}{RGB}{70,165,69}
\definecolor{newpurple}{RGB}{140,69,152}

\definecolor{ibm1}{RGB}{100,143,255}
\definecolor{ibm2}{RGB}{120,94,240}
\definecolor{ibm3}{RGB}{220,38,127}
\definecolor{ibm4}{RGB}{254,97,0}
\definecolor{ibm5}{RGB}{255,176,0}

\pgfdeclarelayer{bg}
\pgfsetlayers{bg,main}

\tikzset{
	arc/.style = {->, semithick, >={[round,sep]Stealth}},
	triangle/.style = {regular polygon, regular polygon sides = 3, inner sep = 0.5pt},
	failure/.style = {triangle, draw, fill = ibm3, minimum size = 0.5cm},
	component/.style = {ellipse, draw, fill = ibm5, minimum size = 0.6cm},
	function/.style = {rectangle, draw, fill = cyan!70, minimum size = 0.6cm},
	action/.style = {regular polygon, regular polygon sides = 5, draw, fill = ibm1, inner sep = 2pt, minimum size = 0.6cm},
	funcvar/.style = {ellipse, draw, fill = gray!15, minimum size = 0.6cm},
}

\journalname{German Journal of Artificial Intelligence}
\begin{document}
\title{Automated Computation of Therapies Using Failure Mode and Effects Analysis in the Medical Domain}
\author{
	Malte Luttermann$^1$ \and
	Edgar Baake$^2$ \and
	Juljan Bouchagiar$^3$ \and
	Benjamin Gebel$^4$ \and
	Philipp Grüning$^5$ \and
	Dilini Manikwadura$^6$ \and
	Franziska Schollemann$^7$ \and
	Elisa Teifke$^8$ \and
 	Philipp Rostalski$^7$ \and
	Ralf Möller$^1$
}
\authorrunning{Luttermann, Baake, Bouchagiar, Gebel, Grüning, Manikwadura, Schollemann, Teifke, Rostalski, Möller}
\institute{
	$^1$Institute of Information Systems, University of Lübeck, Germany \\
	\email{\{luttermann,moeller\}@ifis.uni-luebeck.de} \\
	$^2$Institute of Telematics, University of Lübeck, Germany \\
	\email{baake@itm.uni-luebeck.de} \\
	$^3$Institute for Software Engineering and Programming Languages, University of Lübeck, Germany \\
	\email{juljan.bouchagiar@isp.uni-luebeck.de} \\
	$^4$Department of Infectious Diseases and Microbiology, University Hospital Schleswig-Holstein/ Campus Lübeck, Germany \\
	\email{Benjamin.Gebel@uksh.de} \\
	$^5$Institute for Neuro- and Bioinformatics, University of Lübeck, Germany \\
	\email{ph.gruening@uni-luebeck.de} \\
	$^6$Institute for Molecular Medicine, University of Lübeck, Germany \\
	\email{DiliniTharindika.Manikwadura@uksh.de} \\
	$^7$Institute for Electrical Engineering in Medicine, University of Lübeck, Germany \\
	\email{\{franziska.schollemann,philipp.rostalski\}@uni-luebeck.de} \\
	$^8$Department of Anesthesiology and Intensive Care, University Hospital Schleswig-Holstein/ Campus Lübeck, Germany \\
	\email{Elisa.Teifke@uksh.de}
}

\date{Received: 21 June 2023 / Accepted: 9 September 2023}

\maketitle

\begin{abstract}
	\Ac{fmea} is a systematic approach to identify and analyse potential failures and their effects in a system or process.
    The \ac{fmea} approach, however, requires domain experts to manually analyse the \ac{fmea} model to derive risk-reducing actions that should be applied.
    In this paper, we provide a formal framework to allow for automatic planning and acting in \ac{fmea} models.
    More specifically, we cast the \ac{fmea} model into a \acl{mdp} which can then be solved by existing solvers.
    We show that the \ac{fmea} approach can not only be used to support medical experts during the modelling process but also to automatically derive optimal therapies for the treatment of patients.
\end{abstract}
\acresetall

\section{Introduction}
\Ac{fmea} is a widely used framework to assess the risk of a system or process.
In particular, \ac{fmea} breaks down a system into a hierarchy of components, their functionalities, and potential failures to allow for a systematic analysis of each component or function and their potential failures~\cite{Stamatis2003a}.
Failures are then prioritised according to their potential harm, that is, their severity, their likelihood of occurrence, and their detectability~\cite{Aguirre_FMEA_2021}.
Based on the priorities of the failures, countermeasures against the most critical failures can be developed~\cite{Viscariello_2020}.
The \ac{fmea} approach is an industry standard in the engineering and the manufacturing industry.
The application of \ac{fmea} helps to improve the design of the final product and to detect and reduce the risk of failure~\cite{Aguirre_FMEA_2021,Najafpour_2017}.
\ac{fmea} is already applied during the manufacturing process of medical devices~\cite{liu_use-related_2012} to increase their reliability and helps to satisfy the quality requirements of particular medical processes such as administering drugs to patients~\cite{Adachi2005a}.
Our goal in this paper is to automate planning and acting in \ac{fmea} models.
In particular, we apply \ac{fmea} to the medical domain to automatically derive optimal therapies for individual patients.

Although there is a lot of work on using formal models for diagnosis and treatment of patients, there are, to the best of our knowledge, no applications of \ac{fmea} to model the underlying cause-effect relationships in the human body used in the decision making process of medical experts for diagnosis and treatment.
As the human body is without doubt a highly complex system and all of these formal models must be created manually by domain experts, the \ac{fmea} approach has the potential to support medical experts during the modelling process of cause-effect relationships in the human body.
\Ac{fmea} ensures compliance with the model hierarchy, thereby yielding a structured model approach that helps to deal with the complexity of the model.
Furthermore, as the \ac{fmea} approach in its current form requires a lot of manual work from domain experts even after the \ac{fmea} model has been constructed because an \ac{fmea} model does not provide any automated reasoning capabilities to derive countermeasures (i.e., actions) against potential failures, we propose to cast the \ac{fmea} model into a \ac{mdp}.
An \ac{mdp} provides a mathematical framework to model sequential decision problems in a fully observable, stochastic environment.
In particular, we show how an arbitrary \ac{fmea} model can be transformed into an \ac{mdp} such that existing \ac{mdp} solvers can be applied to automatically derive optimal policies, thereby allowing us to use the transformation from \ac{fmea} model to \ac{mdp} to fully automate the computation of possible countermeasures for potential failures in the \ac{fmea} model.
Further, we demonstrate that a policy obtained from the \ac{mdp} solver can be used to obtain an optimal therapy for an individual patient.

\paragraph{Related work.}
The \ac{fmea} approach is widely applied in various industries such as the automotive industry~\cite{Segismundo2008a}, the aerospace industry~\cite{Yazdi2017a}, and manufacturing industries in general~\cite{Press2003a}.
\Ac{fmea} supports companies during the design phase of their product and is especially prevalent during the design phase of products with special reliability requirements in safety-critical application environments.
In the medical (healthcare) domain, the \ac{fmea} approach is commonly applied during the design phase of medical devices~\cite{Press2018a} as these devices are often used in hospitals and other health-critical environments that must fulfil extraordinary safety requirements.
The administration of drugs to patients in a hospital as well as the evaluation of an automated treatment planning tool for radiation are other safety-critical processes in the medical domain where the \ac{fmea} approach is already being used~\cite{Adachi2005a,kisling_risk_2019}.
However, to the best of our knowledge, the \ac{fmea} approach is not yet applied to model the decision-making process of a medical expert for diagnosis and treatment based on the cause-effect relationships in the human body.
All of the aforementioned industries rely on manual work of domain experts to not only construct the \ac{fmea} model but also to use it to derive actions serving as countermeasures to the potential failures.

\Acp{dss} are approved by many practitioners in the clinical routine for their support in finding the best action while including a growing amount of information and clinical data (\enquote{information overload})~\cite{akbulut_decision_2014,Berner2007a,rudowski_current_1996,Sutton2020a}.
If the standard guidelines are integrated within a \ac{dss}, the outcome is improved~\cite{pelletier_can_2020,rudowski_current_1996}.
Examples of \ac{dss} applications, for instance within the mechanical ventilation domain, are the calculation of initial ventilation parameters~\cite{karbing_open-loop_2018,sward_computerized_2016}, the construction of a weaning protocol for children~\cite{hartmann_interaction_2020}, or the ventilation of patients with acute respiratory distress syndrome~\cite{bottino_decision_1997}.
Nevertheless, \acp{dss} often have a very specific and limited use case wherefore each application requires an individual \ac{dss} to be elaborated and implemented.
In general, there are plenty of works using various mathematical frameworks to establish medical decision support~\cite{VanBaalen2021a}---for example, partially observable \acp{mdp} (a generalisation of \acp{mdp}) are already employed to support in diagnosis and treatment~\cite{Battefeld2022a,Hauskrecht2000a,Zhang2022a}.

\paragraph{Our contributions.}
We first extend the standard definition of an \ac{fmea} model by adding variables (parameters) to functions.
The variables and their qualitative relationships among each other allow us to define a formal semantics of failures and actions in an \ac{fmea} model, i.e., failures indicate that the value of a variable is outside of its normal range and an action restricts the set of possible values for the variables.
Having defined a formal semantics for an \ac{fmea} model, we next show how such a model can be transformed into an \ac{mdp} such that all transition probabilities and rewards can be directly derived from the \ac{fmea} model.
To obtain the possible successor states in the \ac{mdp}, we apply qualitative causal reasoning in the \ac{fmea} model.
The \ac{mdp} can then be solved using existing \ac{mdp} solvers to obtain an optimal policy, which maps each possible state of the system to the best possible action for that particular state.
We present an algorithm to automatically derive the best possible therapy according to the initial \ac{fmea} model for a particular patient using the optimal policy obtained by solving the \ac{mdp}.

\paragraph{Structure of this paper.}
The remaining part of this paper is structured as follows.
\Cref{sec:prelim} introduces the necessary background information for the main part of this paper.
We first define \ac{fmea} models and then introduce \acp{mdp} as a mathematical framework for modelling sequential decision problems in a fully observable, stochastic environment.
Afterwards, in \cref{sec:fmea_mdp}, we show how an \ac{fmea} model can be transformed into an \ac{mdp} which can then be used for automated planning and acting in an \ac{fmea} model.
We introduce qualitative causal reasoning to obtain an algorithm that computes the possible successor states after applying an action in the \ac{mdp}.
\Cref{sec:therapies} introduces an algorithm to compute the best possible therapy for a given patient according to a given \ac{fmea} model.
Finally, we discuss applications and limitations of our approach in \cref{sec:discussion} before we conclude this paper in \cref{sec:conclusion}.

\section{Preliminaries} \label{sec:prelim}
In this section, we introduce the necessary background information for the remainder of this paper.
We begin by formally defining the syntax of an \ac{fmea} model, which we then extend by adding variables to the \ac{fmea} model to obtain an extended \ac{fmea} model for which we can define a formal semantics.
Afterwards, we define \acp{mdp} as a framework to model sequential decision problems in a fully observable, stochastic environment.

\subsection{Failure Mode and Effects Analysis} \label{sec:fmea}
\Ac{fmea} is a systematic approach to identify and analyse potential failures in a system or process~\cite{Stamatis2003a}.
During the \ac{fmea} process, the system is decomposed into its components and functions and for each function, the possible failures are identified.
Every failure is assigned its potential severity, its likelihood of occurrence, and its detectability, which are then combined into a \emph{risk priority number} to assess the risk of the failure.
To be able to apply the \ac{fmea} approach to the medical domain to automatically derive therapies for particular patients, we begin by defining an \ac{fmea} model.
\begin{definition}[\Ac{fmea} Model] \label{def:fmea_model}
	An \emph{\ac{fmea} model} is defined as a tuple $\mathcal{F} = (C, \allowbreak F, \allowbreak E, A, \allowbreak C2C, \allowbreak F2F, \allowbreak E2E, \allowbreak C2F, \allowbreak F2E, \allowbreak A2E, \allowbreak RP, \allowbreak AP)$, where
	\begin{itemize}
		\item $C$ is a finite set of components,
		\item $F$ is a finite set of functions,
		\item $E$ is a finite set of failures,
		\item $A$ is a finite set of actions,
		\item $C2C \subseteq C \times C$ is the component hierarchy,
		\item $F2F \subseteq F \times F$ is the function hierarchy,
		\item $E2E \subseteq E \times E$ is the failure hierarchy,
		\item $C2F \subseteq C \times F$ assigns functions to components,
		\item $F2E \subseteq F \times E$ assigns failures to functions,
		\item $A2E \subseteq A \times E2E$ assigns actions to failure pairs,
		\item $RP \subseteq E \times \{1, \dots, 10\} \times \{1, \dots, 10\} \times \{1, \dots, 10\}$ assigns each failure a severity, occurrence, and detectability value, and
		\item $AP \subseteq A \times \{d,p\}$ specifies the type of each action (\enquote{$d$} for detective or \enquote{$p$} for preventive).
	\end{itemize}
\end{definition}
Note that in its current form, an \ac{fmea} model is only used by domain experts when thinking about potential risks of a system or process, i.e., the model itself does not \enquote{do} anything except for being visually displayed in some kind of graphical user interface.
The hierarchy of components, functions, and failures, induced by $C2C$, $F2F$, and $E2E$, respectively, is constrained to form a connected directed tree (polytree), i.e., a \ac{dag} where any two vertices are connected by exactly one path when replacing the directed edges by undirected edges.
Additionally, each component is restricted to be a sub-component of at most one other component but it is allowed for a component to have multiple sub-components (analogously for functions and failures)---that is, $C2C$, $F2F$, and $E2E$ are left-functional ($1$:$N$) relations.
Further, the relations $C2F$, $F2E$, and $A2E$ are right-total ($N$:$1$), meaning that each function is attached to exactly one component in $C2F$, each failure is attached to exactly one function in $F2E$, and each action is attached to exactly one pair of failure cause and failure effect in $A2E$.
Note that the other way around, there is no restriction, e.g., every component can have arbitrarily many (including zero) functions attached to it.
In general, the idea is that each component supplies one or more functionalities (functions) and each function might go wrong in one or more ways (failures).
Actions are used as a remedy to deal with failures (i.e., functions going wrong).
The first entry in each tuple in $RP$ serves as a key such that there is exactly one tuple contained in $RP$ for each failure $e \in E$, assigning risk parameters (severity, occurrence, and detectability) to $e$.
By $\sev(e)$, $\occ(e)$, and $\detect(e)$ we denote the severity, occurrence, and detectability of $e$, respectively.
We refer to the causes of a failure $e$ by $\causes(e) = \{e' \in E \mid (e', e) \in E2E\}$, and denote its effects by $\effects(e) = \{e' \in E \mid (e, e') \in E2E\}$.
Analogously to $RP$, there is exactly one tuple contained in $AP$ for each action $a \in A$, assigning a type (\enquote{$d$} for detective or \enquote{$p$} for preventive) to $a$.
\begin{example}[\acs{fmea} Model]
	Consider the \ac{fmea} model illustrated in \cref{fig:fmea_example}.
    There are two components \enquote{Perialveolar interstitium} (denoted as $c_1$) and \enquote{Respiratory system} ($c_2$), and the arrow $c_1 \to c_2$ indicates that $c_1$ is a sub-component of $c_2$ (analogously, $f_1$ is a sub-function of $f_2$ and $e_1$ is a cause for $e_2$).
    More specifically, the set of components is given by $C = \{c_1, \allowbreak c_2\}$, the set of functions is given by $F = \{f_1, \allowbreak f_2\}$, the set of failures is given by $E = \{e_1, \allowbreak e_2\}$, and the set of actions is given by $A = \{d_1, \allowbreak p_1\}$.
	The hierarchies for components, functions, and failures are given by $C2C = \{(c_1, c_2)\}$, $F2F = \{(f_1, f_2)\}$, and $E2E = \{(e_1, e_2)\}$, respectively.
	The remaining relations are given by $C2F = \{(c_1, f_1), \allowbreak (c_2, f_2)\}$, $F2E = \{(f_1, e_1), \allowbreak (f_2, e_2)\}$, and $A2E = \{(d_1, \allowbreak (e_1, e_2)), \allowbreak (p_1, \allowbreak (e_1, e_2))\}$.
	Finally, the risk parameters are given by $RP = \{(e_1, 5, 4, 9), \allowbreak (e_2, 7, 5, 9)\}$ and the action parameters are given by $AP = \{(d_1, d), \allowbreak (p_1, p)\}$.
\end{example}
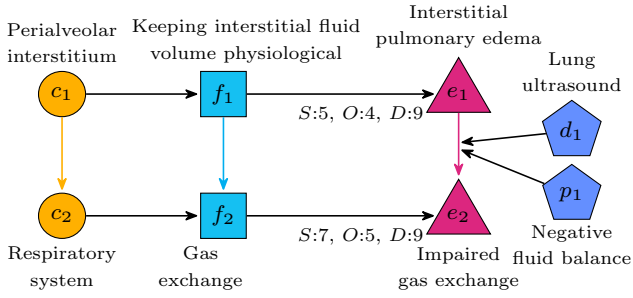
\begin{figure}[t]
	\centering
	\begin{tikzpicture}[
		failure/.append style = {minimum size = 0.4cm},
	]
		\node[component, inner sep = 2pt, label={[align=center]\scriptsize Perialveolar \\ \scriptsize interstitium}] (c1) {$c_1$};
		\node[component, inner sep = 2pt, below = 1cm of c1, label={[align=center, below = 0.6cm]\scriptsize Respiratory \\ \scriptsize system}] (c2) {$c_2$};
	
		\node[function, right = 1.5cm of c1, align = center, label={[align=center, xshift = 0.3cm]\scriptsize Keeping interstitial fluid \\ \scriptsize volume physiological}] (fn1) {$f_1$};
		\node[function, right = 1.5cm of c2, align = center, label={[align=center, below = 0.6cm, xshift = -0.3cm]\scriptsize Gas \\ \scriptsize exchange}] (fn2) {$f_2$};
	
		\node[failure, right = 2.5cm of fn1, align = center, label={[align=center]\scriptsize Interstitial \\ \scriptsize pulmonary edema}] (fail1) {$e_1$};
		\node[failure, right = 2.5cm of fn2, align = center, label={[align=center, below = 0.8cm]\scriptsize Impaired \\ \scriptsize gas exchange}] (fail2) {$e_2$};
	
		\node[below left = -0.2 and 0.1cm of fail1, align = left] (fp1) {\scriptsize $S$:$5$, $O$:$4$, $D$:$9$};
		\node[below left = -0.2 and 0.1cm of fail2, align = left] (fp2) {\scriptsize $S$:$7$, $O$:$5$, $D$:$9$};
	
		\node[action, below right = 0cm and 1cm of fail1, label={[align=center]\scriptsize Lung \\ \scriptsize ultrasound}] (a1) {$d_1$};
		\node[action, below = 0.1cm of a1, label={[align=center, below = 0.7cm]\scriptsize Negative \\ \scriptsize fluid balance}] (a2) {$p_1$};
	
		\draw (c1) edge[arc, ibm5] (c2);
		\draw (c1) edge[arc] (fn1);
		\draw (c2) edge[arc] (fn2);
		\draw (fn1) edge[arc, cyan] (fn2);
		\draw (fn1) edge[arc] (fail1);
		\draw (fn2) edge[arc] (fail2);
		\draw (fail1) edge[arc, ibm3] (fail2);
	
		\path [draw] (a1) edge[arc] ($(fail1) !.4! (fail2)$);
		\path [draw] (a2) edge[arc] ($(fail1) !.45! (fail2)$);
	\end{tikzpicture}
	\caption{An example for an \ac{fmea} model. The components $C$ are given by the circles, the functions $F$ by the rectangles, the failures $E$ by the triangles, and the actions $A$ by the pentagons.
    The hierarchy relations and assignment relations are indicated by the edges between the components, functions, failures, and actions, respectively.
    The name of the actions indicate the action parameters $AP$ (\enquote{$d$} for detective and \enquote{$p$} for preventive) and the risk parameters $RP$ are given by the $S$, $O$, and $D$ values next to the failures.}
	\label{fig:fmea_example}
\end{figure}
As the \ac{fmea} model in its current form is not able to do anything, we next extend an \ac{fmea} model by adding variables (parameters) to functions and afterwards define the semantics of the extended \ac{fmea} model.
\begin{definition}[Extended \ac{fmea} Model] \label{def:ext_fmea_model}
	An \emph{extended \ac{fmea} model} is defined as a tuple $\mathcal{F} = (C, \allowbreak F, \allowbreak E, \allowbreak A, \allowbreak C2C, \allowbreak F2F, \allowbreak E2E, \allowbreak C2F, \allowbreak F2E, \allowbreak A2E, \allowbreak RP, \allowbreak AP,\, \allowbreak pre, \allowbreak post, \linebreak V, \allowbreak F2V, \allowbreak \mathcal G)$, where $C$, $F$, $E$, $A$, $C2C$, $F2F$, $E2E$, $C2F$, $F2E$, $A2E$, $RP$, and $AP$ form an \ac{fmea} model according to \cref{def:fmea_model},
	\begin{itemize}
		\item $pre$ assigns preconditions in form of Boolean expressions to actions,
		\item $post$ assigns postconditions in form of Boolean expressions to actions,
		\item $V$ is a finite set of variables,
		\item $F2V \subseteq F \times V$ assigns variables to functions, and
		\item $\mathcal{G} = (V, E')$ is a directed graph that encodes qualitative relationships between the variables in $V$.
	\end{itemize}
\end{definition}
An action $a \in A$ is called applicable in state $s$ if $s$ satisfies the preconditions of $a$.
Sometimes, an action $a$ might invoke side effects which are captured by the postconditions of $a$.
The relation $F2V$ is left-total ($1$:$N$), meaning that each variable is attached to exactly one function.
We require each function to have at least one variable attached to it but there is no limit of variables being attached to the same function.
The idea behind adding variables to functions is that each function $f(v_1, \dots, v_m) = (v'_1, \dots, v'_k)$ produces an output value for each variable $v'_1, \dots, v'_k$ ($k \geq 1$) attached to $f$.
The purpose of variables is that functions are characterised by the variables attached to them, allowing us to define a formal semantics for failures and actions.
A function $f$ is not required to take any variable as an input (i.e., $m = 0$ is allowed).
If there are input variables $v_1, \dots, v_m$ for a function $f$, $v_1, \dots, v_m$ are outputs of sub-functions of $f$.
Moreover, the associated graph $\mathcal{G} = (V, E')$ models the qualitative relationships between all variables in $V$.
In particular, $E' \subseteq V \times V \times \{+,-,?\}$ is a set of labelled edges, that is, there is an edge $u \overset{\ell}{\to} v$ in $\mathcal{G}$ if $(u, v, \ell) \in E'$.
An edge $u \overset{+}{\to} v$ encodes that increasing $u$ will yield an increase of $v$, $u \overset{-}{\to} v$ means that increasing $u$ will yield a decrease of $v$, and $u \overset{?}{\to} v$ entails that the effect of $u$ on $v$ is unknown.
If a label $\ell$ is irrelevant in a specific context, we omit it and simply write $u \to v$ instead of $u \overset{\ell}{\to} v$.
Vertices that are connected by an edge are called adjacent and are neighbours of each other.
$\Pa(v)$ denotes the set of parents of a variable $v$, i.e., $\Pa(v) = \{u \mid \exists \ell: (u, v, \ell) \in E'\}$ and the children of $v$ are given by $\Ch(v) = \{u \mid \exists \ell: (v, u, \ell) \in E'\}$.
By $\range(v)$ we denote the set of possible values a variable $v \in V$ can take.
For simplicity, we assume that $\range(v) \subseteq \{\tooLow, \normal, \tooHigh\}$ and $\normal \in \range(v)$ for all variables $v$ throughout this paper, i.e., the value of each variable can either be in its normal range or deviate from its normal range in both directions.
Consequently, each failure $e \in E$ has either the form $e := \lc(v_i)$ (i.e., implying that $v_i = \tooLow$) or $e := \rc(v_i)$ (i.e., implying that $v_i = \tooHigh$), where $v_i \in V$ is a variable attached to the function to which $e$ is attached.
For example, if there is a variable $v_i$ called \enquote{body temperature} and a failure $e := \rc(v_i)$ (fever), then $v_i$ can either be assigned the value $\normal$ or $\tooHigh$, while $v_i = \tooHigh$ triggers the failure $e$.
However, different ranges (and hence different failure semantics) are also possible and do not affect our approach to automate planning and acting in an \ac{fmea} model.
\begin{example}[Extended \acs{fmea} Model]
	Take a look at the extended \ac{fmea} model depicted in \cref{fig:fmea_extended_example} which builds on the \ac{fmea} model from \cref{fig:fmea_example}.
    There are now variables $V = \{v_1, \allowbreak v_2\}$ attached to the functions $f_1$ and $f_2$, respectively.
	In particular, the assignments of variables to functions are given by $F2V = \{(f_1, v_1), \allowbreak (f_2, v_2)\}$.
	The qualitative relationships between the variables in $V$ are encoded by the graph $\mathcal G = (V, \allowbreak \{(v_1, v_2, -)\})$.
    To apply $p_1$, there is a precondition that an interstitial pulmonary edema must be detected first, i.e., $pre = \{(p_1, v_1 = \tooHigh)\}$.
	There are no side effects (postconditions) for both of the actions, that is, $post = \emptyset$.
    The model states that too much interstitial fluid volume results in an interstitial pulmonary edema, i.e., $e_1 = \rc(v_1)$, and too little diffusing capacity of the lung impairs the gas exchange, i.e., $e_2 = \lc(v_2)$.
	The edge $v_1 \overset{-}{\to} v_2$ implies that if the interstitial fluid volume is too high, the diffusing capacity of the lung will eventually become too low.
\end{example}
\begin{figure}[t]
	\centering
	\begin{tikzpicture}[
		failure/.append style = {minimum size = 0.4cm},
	]
		\node[component, inner sep = 2pt, label={[align=center]\scriptsize Perialveolar \\ \scriptsize interstitium}] (c1) {$c_1$};
		\node[component, inner sep = 2pt, below = 1cm of c1, label={[align=center, below = 0.6cm]\scriptsize Respiratory \\ \scriptsize system}] (c2) {$c_2$};
	
		\node[function, right = 1.5cm of c1, align = center, label={[align=center, xshift = 0.3cm]\scriptsize Keeping interstitial fluid \\ \scriptsize volume physiological}] (fn1) {$f_1$};
		\node[function, right = 1.5cm of c2, align = center, label={[align=center, below = 0.6cm, xshift = -0.3cm]\scriptsize Gas \\ \scriptsize exchange}] (fn2) {$f_2$};
	
		\node[funcvar, below right = 0.2cm and 0cm of fn1, align = center, inner sep = 2pt, label={[align=center, below right = -0.05cm and 0.3cm]\scriptsize Interstitial \\ \scriptsize fluid volume}] (rv1) {$v_1$};
		\node[funcvar, below right = 0.1cm and 0cm of fn2, align = center, inner sep = 2pt, label={[align=center, below right = 0.3cm and -0.4cm]\scriptsize Diffusing \\ \scriptsize capacity of the lung}] (rv2) {$v_2$};
	
		\node[failure, right = 2.5cm of fn1, align = center, label={[align=center]\scriptsize Interstitial \\ \scriptsize pulmonary edema}] (fail1) {$e_1$};
		\node[failure, right = 2.5cm of fn2, align = center, label={[align=center, below = 0.8cm]\scriptsize Impaired \\ \scriptsize gas exchange}] (fail2) {$e_2$};
	
		\node[below left = -0.2 and 0.1cm of fail1, align = left] (fp1) {\scriptsize $S$:$5$, $O$:$4$, $D$:$9$};
		\node[below left = -0.2 and 0.1cm of fail2, align = left] (fp2) {\scriptsize $S$:$7$, $O$:$5$, $D$:$9$};
	
		\node[action, below right = 0cm and 1cm of fail1, label={[align=center]\scriptsize Lung \\ \scriptsize ultrasound}] (a1) {$d_1$};
		\node[action, below = 0.1cm of a1, label={[align=center, below = 0.7cm]\scriptsize Negative \\ \scriptsize fluid balance}] (a2) {$p_1$};
	
		\draw (c1) edge[arc, ibm5] (c2);
		\draw (c1) edge[arc] (fn1);
		\draw (c2) edge[arc] (fn2);
		\draw (fn1) edge[arc, cyan] (fn2);
		\draw (rv1) edge[arc, gray] node[right] {$-$} (rv2);
		\draw (rv1) edge (fn1.south east);
		\draw (rv2) edge (fn2.south east);
		\draw (fn1) edge[arc] (fail1);
		\draw (fn2) edge[arc] (fail2);
		\draw (fail1) edge[arc, ibm3] (fail2);
	
		\path [draw] (a1) edge[arc] ($(fail1) !.4! (fail2)$);
		\path [draw] (a2) edge[arc] ($(fail1) !.45! (fail2)$);
	\end{tikzpicture}
	\caption{An example for an extended \ac{fmea} model building on the \ac{fmea} model illustrated in \cref{fig:fmea_example}. Now, there is a variable attached to each function in the model and the qualitative relationship between the two variables is encoded by the labelled edge between them.}
	\label{fig:fmea_extended_example}
\end{figure}
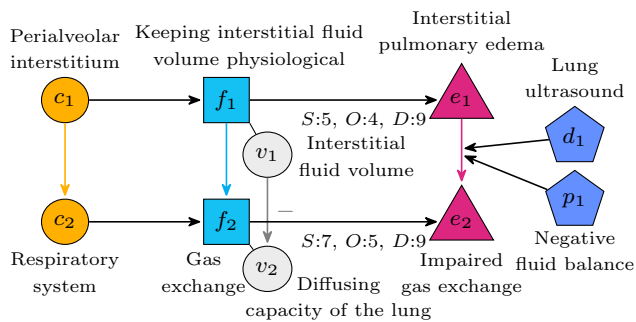
From now on, we focus on extended \ac{fmea} models and simply write \ac{fmea} model instead of extended \ac{fmea} model to refer to an \ac{fmea} model as defined in \cref{def:ext_fmea_model}.
Next, we define the semantics of an \ac{fmea} model whose goal is to assess the risk of a system.
It is important to differentiate between the risk at class level (e.g., the risk of the human body itself) and the risk at instance level (e.g., the risk of a specific human individual).
At class level, we are interested in reducing the risk of the system by changing the model itself, e.g., by adding actions to it.
For example, if a system contains a severe failure that cannot be detected at all, the total risk of the system decreases as soon as a detection action is added to the model to make that failure detectable.
The semantics of an \ac{fmea} model $\mathcal F$ at class level can therefore be defined as
$$
	\risk(\mathcal F) = \max\limits_{e \in E} \phi(\sev(e), \occ(e), \detect(e)),
$$
where $\phi: \{1, \dots, 10\} \times \{1, \dots, 10\} \times \{1, \dots, 10\} \to \{\mathrm{green},\mathrm{orange},\mathrm{red}\}$ is a total function (i.e., $\phi$ is defined for all possible combinations of severity, occurrence, and detectability values) mapping the severity, occurrence, and detectability of a failure to a risk value for that failure.
We require $\{\mathrm{green},\mathrm{orange},\mathrm{red}\}$ to be an ordered set (given in ascending order), i.e., $\max\{\mathrm{green},\mathrm{orange}\} = \mathrm{orange}$ and $\max\{\mathrm{orange},\mathrm{red}\} = \mathrm{red}$.
The risk of an \ac{fmea} model at class level is therefore the risk of the most critical failure in the model.
Other definitions for the semantics of an \ac{fmea} model are possible as well.

For the remaining part of this paper, we focus on assessing the risk of an \ac{fmea} model at instance level.
At instance level, we are interested in determining a sequence of actions that are actually executed to reduce the risk of a particular instance.
For example, in the medical domain, we aim to compute a therapy (sequence of actions) for a particular patient (instance).
Instantiating an \ac{fmea} model for a particular instance yields a state $s$ determined by the possible values each variable $v \in V$ can take.
For example, if it is known that a patient has fever, the variable \enquote{body temperature} is assigned the value \enquote{\tooHigh}.
Applying an action yields a new state and each state is assigned a risk value based on failures that can possibly occur in that state.
The goal is to minimise the risk by performing a sequence of actions to reach a state having a low risk value (i.e., a state corresponding to the patient being healthy).
Before we formally define states and the risk of a state in \cref{sec:fmea_mdp}, we lay the foundations to automate planning and acting in an \ac{fmea} model---that is, to automatically compute the best possible therapy for a specific patient according to the \ac{fmea} model.

\subsection{Markov Decision Processes} \label{sec:mdp}
An \ac{mdp}~\cite{Bellman1957a} is a mathematical framework for modelling a sequential decision problem with discrete time and a fully observable, stochastic environment with a Markovian transition model and additive rewards (i.e., there is a reward in each state and these rewards are added up for the sequence of states that have been visited).
\begin{definition}[\Ac{mdp}]
	We define an \emph{\ac{mdp}} as a tuple $\mathcal{M} = (S, \allowbreak A, \allowbreak s_0, \allowbreak P, \allowbreak R, \allowbreak \gamma)$, where
	\begin{itemize}
		\item $S$ is a finite set of states,
		\item $A$ is a finite set of actions,
		\item $s_0 \in S$ is the initial state,
		\item $P: S \times A \times S \to [0,1]$ is the transition function, i.e., $P(s, a, s')$ yields the probability of transitioning into state $s'$ when taking action $a$ in state $s$,
		\item $R: S \times A \times S \to \mathbb{R}$ is a reward function, i.e., $R(s, a, s')$ is the reward for transitioning to state $s'$ when taking action $a$ in state $s$, and
		\item $\gamma \in [0, 1]$ is the discount factor.
	\end{itemize}
\end{definition}
The discount factor $\gamma$ indicates how much future rewards should be discounted, e.g., $\gamma = 1$ weights all rewards equally while smaller values for $\gamma$ render future rewards less significant.
We write $P(s' \mid s, a)$ to refer to the probability of transitioning to state $s'$ when taking action $a$ in state $s$ ($\sum_{s' \in S} P(s' \mid s, a) = 1$).
\begin{example}[\acs{mdp}]
	\Cref{fig:mdp_example} shows an exemplary \ac{mdp} with states $S = \{s_1, s_2\}$ and actions $A = \{a_1, a_2\}$, depicted as a state-transition system.
	The initial state is $s_1$.
	In each state, both actions can be applied and the transition probabilities are written next to the edges.
	For example, when applying action $a_1$ in state $s_1$, the probability to remain in state $s_1$ is $0.3$ and the probability to transition to state $s_2$ is $0.7$.
	In this particular example, $2 \cdot 2 \cdot 2 = 8$ rewards need to be specified to define the reward function ($R(s_1, a_1, s_1)$, $R(s_1, a_1, s_2)$, and so on).
	We omit the discount factor and the exact specification of the reward function for brevity.
\end{example}
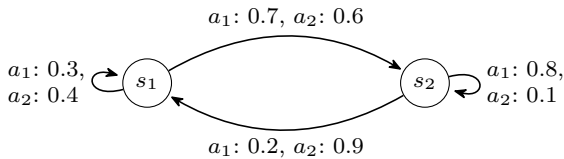
\begin{figure}[t]
	\centering
	\begin{tikzpicture}
		\node[circle, draw] (s1) {$s_1$};
		\node[circle, draw, right = 3cm of s1] (s2) {$s_2$};
		
		\draw[arc] (s1) edge[loop left] node[left, align = left] {$a_1$: $0.3$, \\ $a_2$: $0.4$} (s1);
		\draw[arc] (s1) edge[bend left] node[above] {$a_1$: $0.7$, $a_2$: $0.6$} (s2);
		\draw[arc] (s2) edge[bend left] node[below] {$a_1$: $0.2$, $a_2$: $0.9$} (s1);
		\draw[arc] (s2) edge[loop right] node[right, align = left] {$a_1$: $0.8$, \\ $a_2$: $0.1$} (s2);
	\end{tikzpicture}
	\caption{An example for an \ac{mdp} with states $S = \{s_1, s_2\}$ and actions $A = \{a_1, a_2\}$. The transition probabilities are given by the numbers written next to the edges. Rewards and the discount factor are omitted for brevity.}
	\label{fig:mdp_example}
\end{figure}
A \emph{policy} $\pi: S \to A$ is a total function which maps each state to an action, i.e., $\pi(s)$ returns the action to take in state $s$.
Solving the \ac{mdp} yields an optimal policy $\pi^*$ which maps every state to the best possible action to take in that state.
The optimal sequence of actions is the one having the maximum expected reward, i.e., the optimal policy depends on the choice of the transition probabilities and reward function.
\begin{example}[Policy]
	Consider again the \ac{mdp} depicted in \cref{fig:mdp_example}.
	A possible policy is
	$$
		\pi(s) = \begin{cases}
			a_1 & \text{if } s = s_1 \\
			a_2 & \text{if } s = s_2.
		\end{cases}
	$$
	The optimality of $\pi$ depends on the choice of $R$.
\end{example}
In the next section, we show how an arbitrary \ac{fmea} model can be transformed into an \ac{mdp}, which can then be solved to obtain an optimal policy, allowing us to automatically compute the best possible action to take in a specific state of the model.
We also introduce qualitative causal reasoning to compute the successor states after applying an action in a particular state.

\section{Automated Planning and Acting in \acs{fmea} Models Using \acsp{mdp}} \label{sec:fmea_mdp}
In this section, we show how to automatically compute the best possible sequence of actions for a specific instance of an \ac{fmea} model.
To automatically obtain the best possible sequence of actions, we construct an \ac{mdp} from an \ac{fmea} model.
The \ac{mdp} can then be solved, yielding an optimal policy for decision making.

\subsection{Construction of the Markov Decision Process} \label{sec:mdp_construction}
Given an \ac{fmea} model $\mathcal{F} = (C, \allowbreak F, \allowbreak E, \allowbreak A, \allowbreak C2C, \allowbreak F2F, \allowbreak E2E, \allowbreak C2F, \allowbreak F2E, \allowbreak A2E, \allowbreak RP, \allowbreak AP, \allowbreak pre, \allowbreak post, \allowbreak V, \allowbreak F2V, \allowbreak \mathcal G)$, we construct an \ac{mdp} $\mathcal{M} = (S, \allowbreak A, \allowbreak s_0, \allowbreak P, \allowbreak R, \allowbreak \gamma)$ as follows.

\paragraph{State space.}
The state space $S$ is defined by the possible values each variable in $V = \{v_1, \dots, v_n\}$ can take.
More specifically, $S \subseteq \times_{i = 1}^{n} 2^{\range(v_i)}$ with $2^X$ denoting the power set of $X$ without the empty set.
As there is typically no evidence available in the beginning, the initial state $s_0$ defaults to $s_0 = \langle \range(v_1), \dots, \range(v_n) \rangle$.
However, it is also conceivable to start with a different initial state if there is additional evidence available.
Note that we model the states in a way such that every state is observable, even though there is still uncertainty about the exact values of the variables.
We denote by $\poss_s(v_i)$ the set of possible values for variable $v_i$ in state $s$ (that is, $\poss_s(v_i)$ references the $i$-th component of the state vector $s$).
\begin{example}
	Consider the set of variables $V = \{v_1\}$ with $\range(v_1) = \{\tooLow, \allowbreak \normal, \allowbreak \tooHigh\}$.
	The state space of the \ac{mdp} then consists of seven different states and is given by $S = \{s_1, \dots, s_7\}$ where $s_1 = \langle \{\tooLow\} \rangle$, $s_2 = \langle \{\normal\} \rangle$, $s_3 = \langle \{\tooHigh\} \rangle$, $s_4 = \langle \{\tooLow, \allowbreak \normal\} \rangle$, $s_5 = \langle \{\tooLow, \allowbreak \tooHigh\} \rangle$, $s_6 = \langle \{\normal, \allowbreak \tooHigh\} \rangle$, and $s_7 = \langle \{\tooLow, \allowbreak \normal, \allowbreak \tooHigh\} \rangle$.
	Each state indicates the possible ranges of the variables in $V$, e.g., $s_1$ is the state where it is known that $v_1 = \tooLow$, $s_4$ is the state where it is known that $v_1 \neq \tooHigh$ but there is no information about whether $v_1 = \tooLow$ or $v_1 = \normal$, and so on.
	Without any evidence, the initial state is $s_7$, where $v_1$ may take any value.
\end{example}

\paragraph{Action space and transition probabilities.}
The set of actions $A$ in the \ac{mdp} is directly given by the set of actions from the \ac{fmea} model.
Both the preconditions and the postconditions for the actions are directly transferred to the \ac{mdp} as well, with an additional precondition $\abs{\poss_s(v_i)} > 1$ being added to each detection action $a$ that is used to detect the value of variable $v_i$ in state $s$, i.e., $a$ is only applicable if the value that should be detected is not already known.
If an action $a \in A$ cannot be applied in state $s$ (i.e., its preconditions are not met), we set $P(s' \mid s, a) = 0$ for all successor states $s'$.
Otherwise, let $S' \subseteq S$ be the set of possible successor states after applying action $a$ in state $s$ and let $k = \abs{S'}$.
Note that for a particular instance, there is exactly one successor state when applying action $a$ in state $s$ but in the general \ac{mdp}, all possible successor states that are reachable for any instance need to be considered.
The computation of possible successor states relies on the qualitative relationships encoded in the graph $\mathcal G$ and is described in detail in \cref{sec:succ_states}.
Given the set of possible successor states and the probability $p$ for an action $a$ to be applied successfully, we set $P(s' \mid s, a) = p/k$ for all successor states $s' \in S'$ with $s' \neq s$ (uniform distribution---can also be adjusted if additional information is available).
The probability $p$ of a detection action $a$ (i.e., $(a,d) \in AP$) is given by $\prob(D) = (9-(D-1))/9$ where $D \in \{1, \dots, 10\}$ is the detectability of the failure $e'$ such that $(a, (e', e)) \in A2E$ (i.e., $a$ is attached to the failure pair $(e', e) \in E2E$).
Note that other probability measures are possible as well (e.g., $\prob(D) = (10-D)/10$).
For a prevention action $a$ (i.e., $(a, p) \in AP$), we set $p = \prob(O)$ where $O \in \{1, \dots, 10\}$ is the occurrence of the failure $e'$ such that $(a, (e', e)) \in A2E$.
Moreover, if the application of an action $a$ in state $s$ is not successful, the system remains in state $s$, i.e., $P(s \mid s, a) = 1 - p$, given that $a$ has no postconditions attached to it.
In case an action has postconditions attached to it, they are incorporated into the state regardless of whether the action has been successful, i.e., if an action fails, only the effect of the action itself is not incorporated into the new state whereas the postconditions are.
Finally, for all states $s'' \notin S'$ that are not reachable when applying action $a$ in state $s$, we set $P(s'' \mid s, a) = 0$.
\begin{example}
	Consider an \ac{fmea} model with two failures $e_1 := \rc(v_1)$ and $e_2 := \rc(v_2)$ (i.e., $\range(v_1) = \allowbreak \range(v_2) = \allowbreak \{\normal, \allowbreak \tooHigh\}$) with $(e_1, e_2) \in E2E$ and a prevention action $a$ attached to $(e_1, e_2)$ (without preconditions).
    Let $s = \langle \{\tooHigh\}, \allowbreak \{\normal\} \rangle$.
    Then, the set of possible successor states after applying action $a$ to prevent $e_1$ is given by $S' = \{s'\}$ with $s' = \langle \{\normal\}, \allowbreak \{\normal\} \rangle$.
    Assuming that $a$ always succeeds, applying $a$ sets $\occ(e_1) = 1$ as $a$ prevents $e_1$ from occurring (more details on the exact definition of action semantics are given in \cref{sec:succ_states}).
    In consequence, we obtain $P(s' \mid s, a) = \prob(1) / 1 = (9 - (1 - 1)) / 9 = 1$ as well as $P(s \mid s, a) = 0$, i.e., the action sets $v_1 = \normal$ while $v_2$ is left unchanged.
\end{example}

\paragraph{Reward function.}
The reward for entering the initial state $s_0$ is set to zero, i.e., $R(s, a, s_0) = 0$ for all actions $a$ and states $s$.
For all successor states $s' \neq s_0$, we define the reward $R(s, a, s')$ for going from state $s$ to successor state $s'$ with action $a$ as described below.
Each state $s'$ induces a set of failures that cannot be ruled out in $s'$, e.g., if $\tooLow \in \poss_{s'}(v_i)$, a failure $e := \lc(v_i)$ corresponding to $v_i$ being too low cannot be ruled out (analogously for $\tooHigh$ and failures being $\rc$).
Let $E_{s'}$ denote the set of failures that cannot be ruled out in state $s'$ and let $0 \leq RPN_e \leq 1000$ denote the risk priority number for the failure $e$.
Then, we define the reward for going from any state $s$ to successor state $s'$ with action $a$ as
$$
	R(s, a, s') = \frac{1}{\abs{E_{s'}}} \sum\limits_{e \in E_{s'}} p_e \cdot (1000 - RPN_e),
$$
where $p_e$ is the failure probability for the failure $e$ (if failure probabilities are unknown, $p_e$ can be set to one for every failure $e$).
If $\abs{E_{s'}} = 0$, we set $R(s, a, s') = \infty$.
Note that neither $s$ nor $a$ occur in the right-hand side of the equation, i.e., the reward for changing into state $s'$ does not depend on the previous state $s$ and the performed action $a$.
However, we include both $s$ and $a$ into the left-hand side of the equation to demonstrate that a more fine-grained definition of the reward function is also conceivable if the necessary information is available.
The maximum value for the risk priority number $RPN_e$ is $1000$ and $RPN_e$ is defined as
$$
	RPN_e = \begin{cases}
		\min\limits_{e' \in \causes(e)} RPN_{e'} & \text{if } \causes(e) \neq \emptyset \\
		0 & \text{otherwise},
	\end{cases}
$$
where the risk priority number for each cause $e'$ of $e$ is a product of severity, occurrence, and detectability values.
More specifically, for a cause-effect pair $(e', e)$, we have $RPN_{e'} = \sev(e) \cdot O' \cdot D'$ with $D' = \detect(e')$ if there exists a detection action for $(e', e)$ which is applicable in $s'$ and otherwise $D' = 10$, and $O' = \occ(e')$ if there exists a prevention action attached to $(e', e)$ whose effect is already manifested in $s'$ and otherwise $O' = 10$ (recall that $10$ is the maximum possible number both for the detectability and the occurrence).
The idea is that the risk of each state $s'$ depends on the detectability and the treatability of the failures that cannot be ruled out in $s'$, i.e., if a failure can neither be detected nor treated, it has a high risk priority number assigned to it.
The minimum operator corresponds to a conjunction ($AND$) of failure causes.
Clearly, other operators such as a disjunction ($OR$) of failure causes could be used as well (i.e., $\max$ instead of $\min$ for the computation of $RPN_e$).
The choice of the discount factor $\gamma$ is not part of the transformation from \ac{fmea} model to \ac{mdp} as $\gamma$ is set by the user independent of the \ac{fmea} model.
\begin{example}
    Consider again the \ac{fmea} model consisting of two failures $e_1 := \rc(v_1)$ and $e_2 := \rc(v_2)$ with $(e_1, e_2) \in E2E$.
    It holds that $\range(v_1) = \allowbreak \range(v_2) = \allowbreak \{\normal, \allowbreak \tooHigh\}$ and there is a detection action $a$ attached to $(e_1, e_2)$ (with the default precondition $\abs{\poss_s(v_1)} > 1$ in each state $s$).
    Further, let $\sev(e_1) = 6$, $\occ(e_1) = 5$, $\detect(e_1) = 9$, $\sev(e_2) = 8$, $\occ(e_2) = 4$, and $\detect(e_2) = 9$ and $p_{e_1} = p_{e_2} = 1$ for simplification.
    Then, we have, for example, $R(s, a, \langle \{\tooHigh\}, \allowbreak \{\tooHigh\} \rangle) = \frac{1}{2} \cdot ((1000 - 0) + (1000 - 8 \cdot 10 \cdot 10))$ (note that $D' = 10$ as $a$ is not applicable due to its precondition and $O' = 10$ because there exists no prevention action in this example) and $R(s, a, \langle \{\normal\}, \allowbreak \{\normal\} \rangle) = \infty$ for all states $s$.
\end{example}
Before we continue to present an algorithm to automatically compute the best therapy for a patient using the optimal policy of the \ac{mdp}, we first describe how the possible successor states after the application of an action are computed using qualitative causal reasoning.

\subsection{Computation of Successor States} \label{sec:succ_states}
As failures influence other failures, an action might have an effect not only on the failure $e$ it acts on, but also on other failures that are effected by $e$.
Therefore, we propagate the effect of an action through the failure hierarchy to determine the possible successor states after applying an action $a$ in state $s$.
Before we describe how the effect of an action is propagated through the failure hierarchy using \emph{qualitative causal reasoning}, we first define the semantics of an action.

We begin by defining the semantics of a detection action $a$ in state $s$, assuming that $a$ is applicable in $s$.
Action $a$ is attached to a failure pair $(e', e) \in E2E$ and detects whether the failure cause $e'$ is present.
Let $e' = \lc(v_i)$ ($\rc(v_i)$, respectively).
Then, it holds that $\poss_{s'}(v_i) = \{\tooLow\}$ ($\{\tooHigh\}$) or $\poss_{s'}(v_i) = \{\normal\}$ after transitioning into state $s'$ by applying action $a$ in state $s$---that is, a detection action determines the value of a variable $v_i$ if it succeeds ($a$ always succeeds if $\detect(e') = 1$).
In case $\detect(e') > 1$, $a$ might fail occasionally as we have seen earlier at the construction of the transition probabilities.
Consequently, after applying a detection action $a$ in state $s$, it holds that $\poss_{s'}(v_i) \subseteq \poss_s(v_i)$ for all variables $v_i$ and successor states $s'$ because the detection action reduces the uncertainty about the values of the variables.
Note that in the real world, it might be possible for a detection action $a$ to return an incorrect value (false positive or false negative) if there is a measurement error.
In its current form, the \ac{mdp} does not account for such measurement errors, i.e., the measured value defines the successor state without taking into account that the measured value might be erroneous.

A prevention action $a$ is attached to a failure pair $(e', e) \in E2E$ as well and prevents the failure cause $e'$ from occurring.
More specifically, if $e' = \lc(v_i)$ ($\rc(v_i)$, respectively), then applying action $a$ in state $s$ ensures that $\poss_{s'}(v_i) = \{\normal\}$ after transitioning into state $s'$.
In other words, a prevention action $a$ eliminates the failure cause $e'$ by assigning the value of the corresponding variable $v_i$ to its normal range and hence, we set $\occ(e') = 1$ after applying action $a$ (that is, we assume that the application of $a$ is always successful in preventing $e'$---it is also conceivable that $a$ might fail sometimes which can be modelled by setting $\occ(e')$ to a value greater than one).

Whenever an action is applied successfully, it might affect not only the failure it directly operates on but also other failures in the failure hierarchy.
For example, if we have $(e_1, e_2) \in E2E$ (i.e., $e_1$ causes $e_2$) and an action is applied that prevents $e_1$ from occurring, then $e_2$ cannot occur as well if there are no other causes for $e_2$ other than $e_1$.
Analogously, if a detection action determines that $e_1$ is not present in a particular state $s$, then $e_2$ cannot occur in $s$ as well if there are no other causes for $e_2$ other than $e_1$.
As the presence or absence of a failure might influence the information available about its effects, we employ qualitative reasoning~\cite{Druzdzel1993a} to obtain the possible successor states after applying action $a$ in state $s$.
However, we cannot just apply qualitative reasoning as it is proposed in the literature~\cite{Druzdzel1993a,Wellman1990a} because the propagation does not take into account the causal structure of the failure hierarchy.

Instead, we introduce qualitative \emph{causal} reasoning to propagate changes only along the causal directions of the edges in the failure hierarchy.
In particular, when intervening on a specific failure $e$, all incoming edges of $e$ must be cut off before the changes are propagated through the graph~\cite{Pearl1995a,Pearl2009a}.
\begin{algorithm}[t]
	\SetKwProg{Fn}{function}{}{end}
	\caption{Compute Successor States}
	\label{alg:succ_states}
	\BlankLine
	\Fn{$succ\_states(\mathcal G = (V, E'), a, s)$}{
		$S' \gets \emptyset$\;
        \tcp{$v_i$ is the variable $a$ acts on}
		\ForEach{possible outcome $v_i = r_i$ of $a$}{
            $s' \gets s$\;
            \If{$r_i = \tooLow$}{
                $\sigma \gets$ '$-$'\;
            }
            \ElseIf{$r_i = \tooHigh$}{
                $\sigma \gets$ '$+$'\;
            }
            \Else{
                $\sigma \gets$ '$0$'\;
            }
            $E'' \gets E' \setminus \{(u, v_i, \ell) \mid \exists \ell: (u, v_i, \ell) \in E' \}$\;
			$signs \gets propagate(\mathcal G' = (V, E''), s, v_i, \sigma)$\;
            \ForEach{$(v, \sigma') \in signs$}{
                \If{$\sigma' = \text{'$-$'} \land \tooLow \in \range(v)$}{
                    $s'[v] \gets \{\tooLow\}$\;
                }
                \ElseIf{$\sigma' = \text{'$+$'} \land \tooHigh \in \range(v)$}{
                    $s'[v] \gets \{\tooHigh\}$\;
                }
                \ElseIf{$\sigma' = \text{'$0$'}$}{
                    $s'[v] \gets \{\normal\}$\;
                }
            }
            $push(s', S')$\;
		}
		\Return{$S'$}\;
	}
\end{algorithm}
\Cref{alg:succ_states} depicts the algorithm that is used to compute the set of possible successor states $S'$ when applying an action $a$ in state $s$.
The algorithm considers every possible outcome of the action $a$ in state $s$.
For a prevention action, there is a single outcome per definition, i.e., the value of the corresponding variable is set to a specific value in its range.
The outcome of a detection action, however, is not known when solving the \ac{mdp} as multiple outcomes are possible in practice (e.g., detecting the value of a variable might either yield \enquote{$\tooLow$} or \enquote{$\normal$} and we do not know in advance which of these will be detected for a particular instance).
We denote by $r_i$ the outcome of action $a$ on variable $v_i$, i.e., $a$ acts on $v_i$ and we have $r_i = \normal$ if $a$ is a prevention action, otherwise $r_i$ equals the detected value ($r_i \in \{\tooLow, \allowbreak \normal, \allowbreak \tooHigh\}$).
Thus, it holds that $v_i = r_i$ after applying action $a$, i.e., the $i$-th component of the successor state vector is set to $\{r_i\}$.
We abuse notation and write $s'[v_i]$ instead of $s'[i]$ in the following to refer to the position of a variable $v_i$ in the state vector.
The value $r_i$ of $v_i$ is converted to a sign ('$-$' for \enquote{\tooLow}, '$+$' for \enquote{\tooHigh}, and '$0$' for \enquote{\normal}) which is then propagated through a modified version of the graph $\mathcal G$ (which encodes the qualitative relationships between the variables using edges labelled with signs '$+$', '$-$', and '$?$') where the edges between the parents of $v_i$ and $v_i$ are removed.
The removal of edges between the parents of $v_i$ and $v_i$ in $\mathcal G$ yields a modified graph $\mathcal G'$ which is then used as an input for the qualitative reasoning algorithm illustrated in \cref{alg:propagate_sign} to avoid the propagation of changes against the causal edge direction.

\begin{algorithm}[t]
	\SetKwProg{Fn}{function}{}{end}
	\caption{Qualitative Reasoning (based on the qualitative sign propagation algorithm proposed by Druzdzel and Henrion~\cite{Druzdzel1993a})}
	\label{alg:propagate_sign}
	\BlankLine
	\Fn{$propagate(\mathcal G = (V, E'), s, u, \sigma)$}{
        $signs \gets$ empty dictionary\;
        $vis \gets \emptyset$\;
		\ForEach{$v \in V$}{
			$signs[v] \gets \sign_s(v)$\; \label{line:qr_init}
		}
        $propagate\_rec(\mathcal G, u, u, \sigma, signs, vis)$\; \label{line:qr_first_rec_call}
        \Return{$signs$}\;
	}
	\Fn{$propagate\_rec(\mathcal G, u, v, \sigma, signs, vis)$}{
        $msgs = \{\sigma\}$\; \label{line:qr_msgs_init}
        $signs[v] \gets$ '$0$'\;
        \ForEach{$w \in \Pa(v) \setminus \{u\}$}{
            $\ell \gets$ label of the edge between $w$ and $v$\;
            $msgs \gets msgs \cup \{signs[w] \otimes \ell\}$\;
        }
        $\sigma' \gets signs[v]$\;
        \ForEach{$m \in msgs$}{
            $\sigma' \gets \sigma' \oplus m$\;
        }
		$signs[v] \gets \sigma'$\; \label{line:qr_update_v}
        $vis \gets vis \cup \{v\}$\;
		\ForEach{$w \in \Ch(v)$}{ \label{line:rq_ch_update_loop}
			$\ell \gets$ label of the edge between $v$ and $w$\;
			$m \gets signs[v] \otimes \ell$\;
			\If{$w \notin vis \land signs[w] \neq m$}{
				$propagate\_rec(\mathcal G, v, w, m, signs, vis)$\; \label{line:rq_ch_update_rec_call}
			}
		}
	}
\end{algorithm}

The algorithm for qualitative reasoning builds on the qualitative sign propagation algorithm proposed by Druzdzel and Henrion~\cite{Druzdzel1993a}.
It starts by assigning each variable $v \in V$ a sign in \cref{line:qr_init}, depending on the possible values $v$ can take in state $s$.
In particular, we define
$$
    \sign_s(v) = \begin{cases}
        \text{'$+$'} & \text{if } \poss_s(v) = \{\tooHigh\} \\
        \text{'$-$'} & \text{if } \poss_s(v) = \{\tooLow\} \\
        \text{'$0$'} & \text{if } \poss_s(v) = \{\normal\} \\
        \text{'$?$'} & \text{otherwise}. 
    \end{cases}
$$
The sign of the variable $v_i$ on which the action $a$ has been applied is updated first by calling $propagate\_rec$ with $u = v = v_i$ as parameter in \cref{line:qr_first_rec_call}.
The propagated information (i.e., the signs of the variables) is stored in a dictionary $signs$ and already visited variables are stored in a set $vis$ such that each variable is visited at most once.
During the propagation procedure, we make use of the sign multiplication ($\otimes$) and sign addition ($\oplus$) operators~\cite{Wellman1990a}, which are defined in \cref{tab:sign_operators}.
\begin{table}[t]
    \centering
    \begin{tabular}{c|cccc}
        \toprule
        $\otimes$ & $+$ & $-$ & $0$ & $?$ \\ \midrule
        $+$       & $+$ & $-$ & $0$ & $?$ \\
        $-$       & $-$ & $+$ & $0$ & $?$ \\
        $0$       & $0$ & $0$ & $0$ & $0$ \\
        $?$       & $?$ & $?$ & $0$ & $?$ \\
        \bottomrule
    \end{tabular}
    \qquad
    \begin{tabular}{c|cccc}
        \toprule
        $\oplus$ & $+$ & $-$ & $0$ & $?$ \\ \midrule
        $+$      & $+$ & $?$ & $+$ & $?$ \\
        $-$      & $?$ & $-$ & $-$ & $?$ \\
        $0$      & $+$ & $-$ & $0$ & $?$ \\
        $?$      & $?$ & $?$ & $?$ & $?$ \\
        \bottomrule
    \end{tabular}
    \caption{Definition of the sign multiplication ($\otimes$) and sign addition ($\oplus$) operators according to Wellman~\cite{Wellman1990a}.}
    \label{tab:sign_operators}
\end{table}
In every call of $propagate\_rec$, $u$ propagates its sign to $v$.
During the propagation from $u$ to $v$, all other parents of $v$ are also considered (\cref{line:qr_msgs_init} to \cref{line:qr_update_v}).
More specifically, the algorithm computes a message from all parents of $v$ to $v$ (note that the message from $u$ is already given by $\sigma$) and afterwards uses the sign addition operator to combine the messages of the parents into a new sign for $v$.
After the sign of $v$ has been updated, $v$ propagates its new sign to its children that have not been visited yet by recursively calling $propagate\_rec$ (\cref{line:rq_ch_update_loop} to \cref{line:rq_ch_update_rec_call}).
\begin{figure}[t]
	\centering
	\begin{tikzpicture}
		\node[funcvar, align = center, inner sep = 2pt] (rv1) {$v_1$};
		\node[funcvar, align = center, inner sep = 2pt, below right = 0.75cm and 0.5cm of rv1] (rv3) {$v_3$};
		\node[funcvar, align = center, inner sep = 2pt, above right = 0.75cm and 0.5cm of rv3] (rv2) {$v_2$};
	
		\node[left = 0cm of rv1] (sign_rv1) {$+$};
		\node[right = 0cm of rv2] (sign_rv2) {$+$};
		\node[below = 0cm of rv3] (sign_rv3) {$\text{'$+$'} \oplus \text{'$-$'} = \text{'$?$'}$};
	
		\draw (rv1) edge[arc, gray] node[left] {$+$} (rv3);
		\draw (rv2) edge[arc, gray] node[right] {$-$} (rv3);
	
		\draw (sign_rv1) edge[arc, dashed, gray!50, bend right] node[black, left, xshift = -0.1cm] {$\text{'$+$'} \otimes \text{'$+$'} = \text{'$+$'}$} (rv3);
		\draw (sign_rv2) edge[arc, dashed, gray!50, bend left] node[black, right, xshift = 0.1cm] {$\text{'$+$'} \otimes \text{'$-$'} = \text{'$-$'}$} (rv3);
	\end{tikzpicture}
	\caption{An example for propagating the sign of $v_1$ to $v_3$. During the propagation, all other parents of $v_3$ (here only $v_2$) are also taken into account. Both $v_1$ and $v_2$ are assigned the sign '$+$' and therefore $v_3$ receives the two messages $\text{'$+$'} \otimes \text{'$+$'} = \text{'$+$'}$ and $\text{'$+$'} \otimes \text{'$-$'} = \text{'$-$'}$, which are then combined using the sign addition operator to obtain $\text{'$+$'} \oplus \text{'$-$'} = \text{'$?$'}$ as a new sign for $v_3$.}
	\label{fig:qr_example_small}
\end{figure}
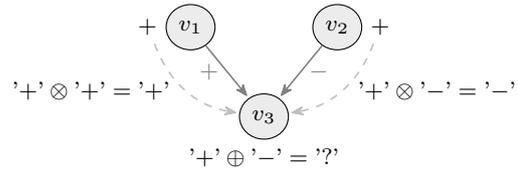
\begin{example}
    Consider the graph shown in \cref{fig:qr_example_small} with edges $v_1 \overset{+}{\to} v_3$ and $v_2 \overset{-}{\to} v_3$ and assume that $v_1$ is assigned the sign '$+$' which is then propagated to its children, i.e., to $v_3$.
    If we also have evidence for $v_2$ suggesting that its value is \enquote{$\tooHigh$} (i.e., the sign of $v_2$ is '$+$'), however, we cannot assign $v_3$ the result of the propagation from $v_1$ to $v_3$ (i.e., $\text{'$+$'} \otimes \text{'$+$'} = \text{'$+$'}$) because there is another influence of $v_2$ on $v_3$ (i.e., $\text{'$+$'} \otimes \text{'$-$'} = \text{'$-$'}$).
    Consequently, we know that $v_3$ both increases (due to $v_1$) and decreases (due to $v_2$) at the same time and we cannot infer whether $v_3$ will eventually become \enquote{$\tooLow$} or \enquote{$\tooHigh$} in this situation.
    The algorithm therefore uses sign addition on all messages from the parents of $v_3$, such that the new sign for $v_3$ is $\text{'$+$'} \oplus \text{'$-$'} = \text{'$?$'}$.
\end{example}
After the propagation of signs is finished, the new signs are returned and converted back to values in $\{\tooLow, \allowbreak \normal, \allowbreak \tooHigh\}$ (in \cref{alg:succ_states}).
The new state $s'$ obtained from the propagation is then added to the list of possible successor states.
Performing qualitative causal reasoning instead of just setting the value of the variable $v_i$ decreases the number of reachable states, i.e., states that contain inconsistent information cannot be reached and hence the state space is smaller than it would be without the propagation of information.
\begin{figure}[t]
	\centering
	\begin{tikzpicture}[
		failure/.append style = {minimum size = 0.4cm},
		action/.append style = {minimum size = 0.8cm},
	]
		\node[component, inner sep = 2pt] (c1) {$c_1$};
		\node[component, inner sep = 2pt, below = 1cm of c1] (c2) {$c_2$};
		\node[component, inner sep = 2pt, below = 1cm of c2] (c3) {$c_3$};
	
		\node[function, right = 1.5cm of c1, align = center] (fn1) {$f_1$};
		\node[function, right = 1.5cm of c2, align = center] (fn2) {$f_2$};
		\node[function, right = 1.5cm of c3, align = center] (fn3) {$f_3$};
	
		\node[funcvar, below right = 0.2cm and 0cm of fn1, align = center, inner sep = 2pt] (rv1) {$v_1$};
		\node[funcvar, below right = 0.1cm and 0cm of fn2, align = center, inner sep = 2pt] (rv2) {$v_2$};
		\node[funcvar, below right = 0.1cm and 0cm of fn3, align = center, inner sep = 2pt] (rv3) {$v_3$};
	
		\node[above right = -0.1cm and -0.1cm of rv1] (sign_rv1) {$+$};
		\node[above right = -0.1cm and -0.1cm of rv2] (sign_rv2) {$+$};
		\node[above right = -0.1cm and -0.1cm of rv3] (sign_rv3) {$+$};
	
		\node[failure, right = 2.5cm of fn1, align = center] (fail1) {$e_1$};
		\node[failure, right = 2.5cm of fn2, align = center] (fail2) {$e_2$};
		\node[failure, right = 2.5cm of fn3, align = center] (fail3) {$e_3$};
	
		\node[action, below right = 0.4cm and 1cm of fail2] (a1) {$a$};
	
		\draw (c1) edge[arc, ibm5] (c2);
		\draw (c2) edge[arc, ibm5] (c3);
		\draw (c1) edge[arc] (fn1);
		\draw (c2) edge[arc] (fn2);
		\draw (c3) edge[arc] (fn3);
		\draw (fn1) edge[arc, cyan] (fn2);
		\draw (fn2) edge[arc, cyan] (fn3);
		\draw (rv1) edge[arc, gray] node[right, yshift = 0.05cm] {$+$} (rv2);
		\draw (rv2) edge[arc, gray] node[right, yshift = 0.05cm] {$+$} (rv3);
		\draw (rv1) edge (fn1.south east);
		\draw (rv2) edge (fn2.south east);
		\draw (rv3) edge (fn3.south east);
		\draw (fn1) edge[arc] (fail1);
		\draw (fn2) edge[arc] (fail2);
		\draw (fn3) edge[arc] (fail3);
		\draw (fail1) edge[arc, ibm3] (fail2);
		\draw (fail2) edge[arc, ibm3] (fail3);
	
		\path [draw] (a1) edge[arc] ($(fail2) !.4! (fail3)$);
	\end{tikzpicture}
	\caption{A visualisation of an \ac{fmea} model with current state $s = \langle \{\tooHigh\}, \allowbreak \{\tooHigh\}, \allowbreak \{\tooHigh\} \rangle$ (the variable assignments $v_1 = v_2 = v_3 = \tooHigh$ are indicated by the '$+$' signs next to the variables). Action $a$ is a preventive action for $e_2$, i.e., applying $a$ sets $v_2 = \normal$. The effect of $a$ is then propagated according to the failure hierarchy such that the successor state after applying $a$ in $s$ is $s' = \langle \{\tooHigh\}, \allowbreak \{\normal\}, \allowbreak \{\normal\} \rangle$.}
	\label{fig:qr_example_detailed}
\end{figure}
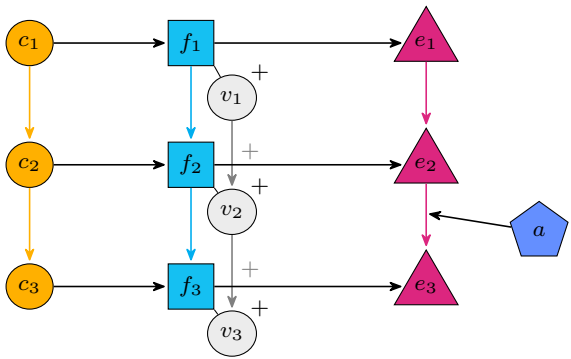
\begin{example}
    For a more comprehensive example of the qualitative causal reasoning algorithm to compute possible successor states, take a look at the \ac{fmea} model depicted in \cref{fig:qr_example_detailed}.
    The \ac{fmea} model contains three failures $e_1 := \rc(v_1)$, $e_2 := \rc(v_2)$, and $e_3 := \rc(v_3)$ with failure hierarchy $e_1 \to e_2 \to e_3$, i.e., $(e_1, e_2) \in E2E$ and $(e_2, e_3) \in E2E$.
    It holds that $\range(v_1) = \allowbreak \range(v_2) = \allowbreak \range(v_3) = \allowbreak \{\normal, \allowbreak \tooHigh\}$.
    Further, let $a$ be a prevention action attached to $(e_2, e_3)$ and $\mathcal G = (\{v_1, v_2, v_3\}, \allowbreak \{(v_1, v_2, +), \allowbreak (v_2, v_3, +)\})$ be the corresponding graph encoding the qualitative relationships between $v_1$, $v_2$, and $v_3$.
    When applying action $a$ in state $s = \langle \{\tooHigh\}, \allowbreak \{\tooHigh\}, \allowbreak \{\tooHigh\} \rangle$, the successor states are computed as follows.
    As $a$ sets $v_2 = \normal$, $\sigma = \text{'$0$'}$ is propagated starting from $v_2$.
    During initialisation in \cref{line:qr_init}, the algorithm assigns $signs[v_1] = signs[v_2] = signs[v_3] = \text{'$+$'}$ (because $v_1 = v_2 = v_3 = \tooHigh$ in $s$).
    In the first call of $propagate\_rec$, it holds that $u = v = v_2$ and hence there are no parents of $v$ (because the ingoing edges of $v_2$ have been removed before calling $propagate$).
    Hence, after initially setting $signs[v_2] = \text{'$0$'}$, the first update being made is $signs[v_2] = signs[v_2] \oplus \sigma = \text{'$0$'} \oplus \text{'$0$'} = \text{'$0$'}$.
    Then, $v_2$ is marked as visited and in the next call of $propagate\_rec$, it holds that $u = v_2$, $v = v_3$, and $\sigma = signs[v_2] \otimes \text{'$+$'} = \text{'$0$'}$.
    As $v_3$ has no other parents apart from $v_2$, the next update being made after initially setting $signs[v_3] = \text{'$0$'}$ is $signs[v_3] = signs[v_3] \oplus \sigma = \text{'$0$'} \oplus \text{'$0$'} = \text{'$0$'}$.
    Afterwards, $v_3$ is marked as visited and the propagation algorithm terminates as there are no children for $v_3$.
    Finally, the signs of the variables are translated back to values in their range such that the new state is $s' = \langle \{\tooHigh\}, \allowbreak \{\normal\}, \allowbreak \{\normal\} \rangle$.
    As $v_2 = \normal$ is the only possible outcome of $a$, there are no other successor states apart from $s'$.
    Note that the value of $v_1$ is left unchanged as the prevention action performed on $v_2$ has no effect on $v_1$.
\end{example}
If additional information about quantitative relationships between variables is available, it is also possible to use quantitative causal reasoning~\cite{Pearl2009a} instead of qualitative causal reasoning.
Before we show how solving the \ac{mdp} of an \ac{fmea} model yields optimal therapies for patients in the medical domain, we note that it is also conceivable to employ a partially observable \ac{mdp}~\cite{Kaelbling1998a,Astrom1965a} instead of an \ac{mdp} to formalise the \ac{fmea} model.
However, the transformation from \ac{fmea} model to partially observable \ac{mdp} is not immediately clear and dealing with partially observable \acp{mdp} is far more complex than dealing with \acp{mdp}, which is a major drawback especially in the medical domain where acquiring the observation probabilities required by a partially observable \ac{mdp} is a highly difficult task (as these probabilities are usually not known).

Next, we give a full algorithm taking as input an \ac{fmea} model and patient data of a specific patient to return an optimal therapy for that patient.

\section{Automated Computation of Optimal Therapies in the Medical Domain} \label{sec:therapies}
Solving the \ac{mdp} from \cref{sec:fmea_mdp} yields an optimal policy $\pi^*$ which maps every state to an optimal action.
The optimal policy $\pi^*$ can then be used to compute therapies for patients.
Executing an action $a = \pi^*(s)$ in state $s$ for an individual patient yields a specific successor state $s'$ for which $\pi^*$ also returns the best possible action to take.
Thus, the optimal policy $\pi^*$ directly corresponds to an optimal therapy.
By adding a goal state to the \ac{mdp} or a threshold on the reward, we can formulate an algorithm that computes the optimal therapy according to a given \ac{fmea} model.

\begin{algorithm}[t]
	\SetKwProg{Fn}{function}{}{end}
	\caption{Compute Optimal Therapy}
	\label{alg:therapy}
	\BlankLine
	\Fn{optimal\_therapy$(\mathcal F, s_0, s_g, \theta, D)$}{
		$\mathcal M \gets fmea\_to\_mdp(\mathcal F, s_0)$\;
		$\pi^* \gets solve\_mdp(\mathcal M)$\;
		$therapy \gets [\,]$\;
		$s' \gets s_0$\;
		\Repeat{$s' = s_g \lor R(s, a, s') > \theta$}{
			$s \gets s'$\;
			$a \gets \pi^*(s)$\;
			$push(a, therapy)$\;
			$s' \gets a(s, D)$\;
		}
		\Return{$therapy$}\;
	}
\end{algorithm}

\Cref{alg:therapy} outlines how to compute the optimal therapy for a specific patient according to a given \ac{fmea} model $\mathcal F$.
The initial state $s_0$ is given by the available evidence for the patient.
First, the algorithm transforms $\mathcal F$ into an \ac{mdp} $\mathcal M$ and then solves $\mathcal M$ to obtain the optimal policy $\pi^*$.
The algorithm then iteratively applies the optimal policy to the current state $s$ until either the goal state $s_g$ (it is also conceivable to have a set of goal states instead of a single goal state) is reached or the reward $R(s, a, s')$ reaches a user-defined threshold $\theta$.
All actions that have been applied as part of the optimal policy are appended to the resulting therapy.
Whenever an action $a$ is applied, the patient data $D$ are taken into account to determine the unique successor state $s'$ (the patient data contains the information about the exact result of the applied action, e.g., the outcome of a detection action).
\begin{example} \label{example:therapy}
	Take a look again the example shown in \cref{fig:fmea_extended_example}.
	The model states that too much interstitial fluid volume results in an interstitial pulmonary edema, i.e., $e_1 = \rc(v_1)$, and too little diffusing capacity of the lung impairs the gas exchange, i.e., $e_2 = \lc(v_2)$.
	In particular, it holds that $\range(v_1) = \{\normal, \allowbreak \tooHigh\}$ and $\range(v_2) = \{\normal, \allowbreak \tooLow\}$ and the edge $v_1 \overset{-}{\to} v_2$ implies that if $v_1$ increases, it causes $v_2$ to decrease, i.e., if the interstitial fluid volume is too high, the diffusing capacity of the lung will eventually become too low.
	Moreover, we have $pre = \{(p_1, \allowbreak v_1 = \tooHigh)\}$, i.e., the action $p_1$ can only be applied if an interstitial pulmonary edema is detected.
    For the sake of this example, let $\sev(e_1) = 5$, $\occ(e_1) = 4$, $\detect(e_1) = 9$, $\sev(e_2) = 7$, $\occ(e_2) = 5$, $\detect(e_2) = 9$, $p_{e_1} = 0.4$, and $p_{e_2} = 0.5$.
    The corresponding \ac{mdp} to this \ac{fmea} model consists of the action space $A = \{d_1, p_1\}$ and the state space $S = \{s_1, \dots, s_9\}$, where
    \begin{itemize}
        \item $s_1 = \allowbreak (\{\normal\},\allowbreak \{\normal\})$,
        \item $s_2 = \allowbreak (\{\normal\},\allowbreak \{\tooLow\})$,
        \item $s_3 = \allowbreak (\{\tooHigh\},\allowbreak \{\normal\})$,
        \item $s_4 = \allowbreak (\{\tooHigh\},\allowbreak \{\tooLow\})$,
        \item $s_5 = \allowbreak (\{\normal, \allowbreak \tooHigh\},\allowbreak \{\normal\})$,
        \item $s_6 = \allowbreak (\{\normal, \allowbreak \tooHigh\},\allowbreak \{\tooLow\})$,
        \item $s_7 = \allowbreak (\{\normal\},\allowbreak \{\normal, \tooLow\})$,
        \item $s_8 = \allowbreak (\{\tooHigh\},\allowbreak \{\normal, \tooLow\})$, and
        \item $s_9 = \allowbreak (\{\normal, \allowbreak \tooHigh\},\allowbreak \{\normal, \allowbreak \tooLow\})$.
    \end{itemize}
	The initial state for a patient without evidence is $s_9$ and the goal state in this example is $s_1$.
    Solving the \ac{mdp} corresponding to the given \ac{fmea} model then yields a policy $\pi^*$ that returns appropriate actions for each state, e.g., $\pi^*(s_4) = p_1$ such that the goal state is reached immediately after applying $p_1$ in $s_4$ (due to the propagation of the effect of $p_1$).
    If we consider a patient arriving at a hospital who has an interstitial pulmonary edema and thus an impaired gas exchange (but this diagnosis is not known beforehand), the optimal therapy computed by \cref{alg:therapy} would be $\langle d_1, p_1 \rangle$ (because the patient data $D$ tells us that applying $d_1$ in state $s_9$ results in a transition to state $s_4$).
    In other words, the recommended therapy would be to first apply the detection action $d_1$ (which then finds out about the patient's interstitial pulmonary edema) and afterwards to apply the prevention action $p_1$ (whose preconditions are then satisfied) to treat the disease accordingly.
\end{example}
Before we conclude this paper, we discuss further applications and limitations of our proposed approach.

\section{Discussion} \label{sec:discussion}
The general approach of transforming an \ac{fmea} model into an \ac{mdp} to automatically compute the best sequence of actions to reduce the risk as much as possible is obviously not restricted to the medical domain.
Hence, industries such as the automotive industry, the aerospace industry, and manufacturing industries in general, which commonly apply the \ac{fmea} approach, can also vastly benefit from the automatic planning and acting capabilities provided by the \ac{mdp}.

However, the presented approach to transform an \ac{fmea} model into an \ac{mdp} clearly has its own limitations and can be further refined, e.g., by integrating different ranges of variables and hence additional failures having semantics that are different from $\lc(v_i)$ and $\rc(v_i)$, by adding costs to actions, by handling erroneous measurements obtained from detection actions, and so on.
It is also possible to incorporate a probability distribution over the variables in $V$ to allow for probabilistic (quantitative) causal inference~\cite{Pearl2009a} instead of merely using qualitative causal inference.
Obtaining more fine-grained \ac{fmea} models, however, is a serious challenge in the medical domain---and most likely also in other domains---as obtaining the additionally required information involves a lot of effort.
Another limitation of the \ac{mdp} is its scalability.
As we have seen in \cref{example:therapy}, the state space of the \ac{mdp} becomes quite large even for a small \ac{fmea} model.
Even though the propagation of action effects during the computation of successor states results in many states not being reachable at all (and hence they could be omitted from the \ac{mdp} after an initial reachability check), the size of the state space is still a limitation when it comes to solving the \ac{mdp} for large \ac{fmea} models.
To encounter the scalability problem induced by large state spaces, reinforcement learning~\cite{Sutton2018a} might be applied as a remedy.
Similar to the AlphaZero program~\cite{Silver2018a} (which has been developed to master the games of chess, shogi, and go where the state spaces are also huge), one could use reinforcement learning to learn an approximation of the optimal policy.
The idea is to sample an initial state and random sequences of actions for which then the reward of the resulting state is used as a measure of quality for that particular action sequence.
By repeating the sampling procedure for various initial states and action sequences, an approximation of the optimal policy can be obtained.
The development of such a reinforcement learning approach, however, is out of the scope of this paper and hence an interesting direction for future work.

Before we conclude this paper, we give an outlook on possibilities to augment \acp{llm} with formalised domain knowledge represented in formal models such as \ac{fmea} models and their corresponding \ac{mdp}.
We believe that formal models can be used to generate training data for the fine-tuning step of an \ac{llm} by sampling the model.
The \ac{mdp} allows us to generate training data for \acp{llm} by computing therapies for a lot of different initial states, thresholds, and patient data.
Given the generated data, the computed therapies can be verbalised (i.e., translated to natural language) and afterwards, the verbalised data can be used to fine-tune a pre-trained \ac{llm}.
By integrating the knowledge of domain experts represented in the formal model into the \ac{llm}, the \ac{llm} might produce less hallucinations.
Furthermore, it is conceivable to use a formal model again to validate the output of the \ac{llm} by translating the output of the \ac{llm} into the syntax of the formal model and then using the formal model to check whether the input-output pair of the \ac{llm} matches the computed optimal therapy.

\section{Conclusion} \label{sec:conclusion}
We present a formal framework to conduct automated planning and acting in \ac{fmea} models.
In particular, we apply \ac{fmea} to the medical domain and transform the resulting \ac{fmea} model into an \ac{mdp} to automatically compute optimal therapies for individual patients.
Further, we introduce qualitative causal reasoning to compute the successor states in the \ac{mdp} after applying an action, yielding a fully automated algorithm to compute a therapy for a particular patient.

Future work includes the application of reinforcement learning to encounter the state space explosion of the \ac{mdp}, as well as the augmentation of general \acp{llm} with formalised domain knowledge.

\section*{Acknowledgements}
This work is funded by the Medical Cause and Effects Analysis (MCEA) project.
The authors also thank the anonymous reviewers for their insightful comments.
This version of the article has been accepted for publication, after peer review, but is not the Version of Record and does not reflect post-acceptance improvements, or any corrections. The Version of Record is available online at: \url{https://doi.org/10.1007/s13218-023-00810-z}.

\bibliographystyle{splncs04}
\bibliography{references.bib}
\end{document}